\documentclass[10pt,twocolumn,letterpaper]{article}
\pdfoutput=1
\usepackage[hyphens]{url}
\usepackage{graphicx}
\usepackage{booktabs}
\usepackage[dvipsnames, svgnames, x11names, table]{xcolor}
\usepackage[most]{tcolorbox}

\usepackage{amssymb}
\usepackage{amsmath}
\usepackage{float}
\usepackage{multirow} 

\usepackage[review]{cvpr}      

\newcommand{\calF}{\mathcal{F}}
\newcommand{\calG}{\mathcal{G}}

\newcommand{\R}{\mathbb{R}}
\providecommand{\bydef}{\overset{\text{def}}{=}}

\definecolor{cvprblue}{rgb}{0.21,0.49,0.74}
\usepackage[pagebackref,breaklinks,colorlinks,allcolors=cvprblue]{hyperref}

\title{AI Tailoring: Evaluating Influence of Image Features on Fashion Product Popularity}

\author{
Xiaomin Li\thanks{Equal contribution.}\\
John A. Paulson School of Engineering and Applied Sciences, Harvard University\\
{\tt\small xiaominli@g.harvard.edu}
\and
Junyi Sha\footnotemark[1]\\
Center for Computational Science \& Engineering, Massachusetts Institute of Technology\\
{\tt\small jsha@mit.edu}
}
\date{} 

\begin{document}
\maketitle
\begin{abstract}
Identifying key product features that influence consumer preferences is essential in the fashion industry. In this study, we introduce a robust methodology to ascertain the most impactful features in fashion product images, utilizing past market sales data. First, we propose the metric called ``influence score" to quantitatively assess the importance of product features. Then we develop a forecasting model, the Fashion Demand Predictor (FDP), which integrates Transformer-based models and Random Forest to predict market popularity based on product images. We employ image-editing diffusion models to modify these images and perform an ablation study, which validates the impact of the highest and lowest-scoring features on the model’s popularity predictions. Additionally, we further validate these results through surveys that gather human rankings of preferences, confirming the accuracy of the FDP model’s predictions and the efficacy of our method in identifying influential features. Notably, products enhanced with ``good" features show marked improvements in predicted popularity over their modified counterparts. Our approach develops a fully automated and systematic framework for fashion image analysis that provides valuable guidance for downstream tasks such as fashion product design and marketing strategy development.

\end{abstract}    
\section{Introduction} \label{sec:Intro}
In the domain of image feature analysis, contemporary approaches increasingly rely on deep learning techniques to extract and interpret complex patterns within image data. Pioneering models such as Convolutional Neural Networks (CNNs) \cite{lecun1989backpropagation}, Vision Transformers (ViT) \cite{dosovitskiy2020image}, and Diffusion Models \cite{sohl2015deep, song2020improved} have demonstrated efficacy across diverse applications, including medical imaging, retail, and e-commerce \cite{zhang2022makes, medus2021hyperspectral, chen2021vit, goenka2022fashionvlp, sultan1990meta, liu2023generative}. These models can be applied as tools for image editing tasks, significantly enhancing image customization for specific needs. For instance, DALL-E \cite{ramesh2021zero} and DALL-E 2 \cite{ramesh2022hierarchical} generate imaginative images from textual descriptions, while StyleGAN enables realistic facial feature manipulations \cite{karras2019style}, and StyleCLIP merges StyleGAN’s capabilities with text-driven controls for detailed edits \cite{patashnik2021styleclip}.

Particularly, for fashion product images, identifying features that influence consumer preferences is crucial. Nonetheless, current methodologies often rely heavily on human input, requiring designers to provide text instructions \cite{pernuvs2025fice, wang2024texfit} or detailed sketches \cite{cui2018fashiongan, yan2022toward}, which introduces subjectivity and potential biases. This highlights the importance of developing objective, automated systems that can accurately identify and modify key features, enhancing both the precision and reliability of fashion design processes.

\begin{figure*}[ht]
    \centering
    \includegraphics[width=\textwidth]{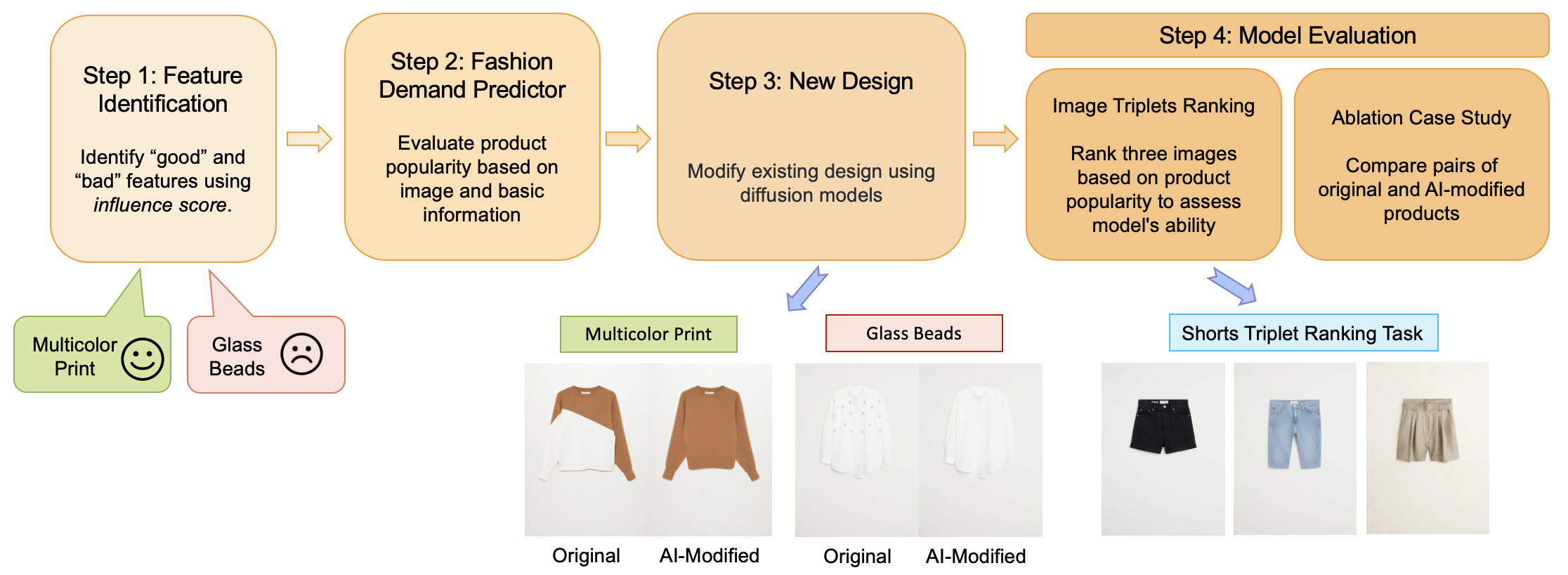}  %
    \caption{Pipeline of our methodology}
    \label{fig:pipeline}
\end{figure*}

In this paper, we introduce a novel method aimed at identifying and evaluating the most influential features in fashion product images, thus guiding the enhancement of design decisions based on their impact on product popularity. We first introduce the ``influence score", aggregated from the popularity data of related products, to quantify the importance of each design feature. This score is adjusted according to the frequency of features observed across a collection of images. Then we train a forecasting model which we call the \textit{Fashion Demand Predictor (FDP)} using real sales data to forecast product popularity, as an increase in consumer ratings has been demonstrated to correspond with higher sales\cite{moen2017online}. To validate our model, we conduct various experiments using real data adapted from a European fast fashion company. First, we assess the effectiveness of the FDP using a held-out test dataset and using human preference data collected through a survey. Secondly, we perform ablation studies to validate our feature influence measure, comparing pairs of original and AI-modified (using  \textit{InstructPix2pix-Distill}\cite{instruct_pix2pix_distill} and \textit{Adobe Firefly Image 3} \cite{adobe2024firefly}) images to analyze the impact of specific features on product popularity, thereby ensuring consistency between model predictions and human assessments. Moreover, we use multi-modal large language models (\textit{Llava-v1.5-7b} \cite{liu2024visual}) as our baseline for comparison. Our approach provides actionable insights for feature identification in fashion product images, essential for targeted image editing and fashion design tasks. The detailed pipeline of our methodology is illustrated in Figure~\ref{fig:pipeline}.

In summary, our contributions are the following: (1) We developed a popularity forecasting model that predicts the popularity of fashion products based on their image and basic description. (2) We propose a quantitative measure to evaluate the influence and importance of distinct design features, where high scores indicate ``good" features and low scores indicate ``bad" features. (3) We provide comprehensive experiments and evaluation metrics to validate the performance of our forecasting model and the effectiveness of the feature influence score, employing ablation studies and human surveys for robust validation.

\section{Related Work} \label{sec:RelatedWork}

\noindent \textbf{Attribute/Feature Extraction}\hspace{0.1cm}
Image feature extraction remains a fundamental task in computer vision, employing a spectrum of techniques from traditional algorithms, such as edge detection and texture analysis \cite{ziou1998edge, tuceryan1993texture}, to more advanced neural network-based methods. Notably, deep learning models like CNNs \cite{lecun2015deep}, Transformers \cite{vaswani2017attention}, and Vision Transformers (ViT) \cite{DBLP:journals/corr/abs-2010-11929} have revolutionized this domain by automatically learning complex features. In the context of fashion images, both conventional approaches \cite{chao2009framework, yang2011real, acsirouglu2019smart} and deep learning techniques \cite{bracher2016fashion, simo2016fashion, lee2017style2vec, zhang2018representative} have been extensively utilized for feature recognition and extraction. In particular, CNN is frequently applied in this field \cite{eshwar2016apparel, li2019retrieving, vittayakorn2016automatic, deng2018research, tu2010intelligent, ma2019and, jia2020fashionpedia}. In our dataset, feature extraction is simplified by using a composite string of descriptors for each fashion product, thereby eliminating the need to design the concepts to describe the image features. Instead, we focus on analyzing and evaluating the importance of features. Our forecasting model leverages FashionCLIP \cite{chia2022fashionclip}, a transformer-based approach that converts image features into latent embeddings.

\vspace{0.2cm}
\noindent \textbf{Fashion Image Generation and Editing}\hspace{0.1cm}
Fashion image generation has been progressively studied with the advent of text-to-image synthesis, especially pertinent to fashion contexts \cite{ma2017pose, esser2018variational, han2018viton, yi2019text}. Generative adversarial networks (GANs) are prominently utilized for generating new designs from sketches \cite{yan2022toward} and for enabling the creation of similar designs through Attribute-GAN, which combines given designs with textual attributes \cite{liu2019toward}. The integration of advanced models such as transformers \cite{vaswani2017attention, ramesh2022hierarchical, radford2021learning, reddy2021dall, ramesh2021zero, patashnik2021styleclip, sbai2018design, zhang2021m6} and diffusion models \cite{ho2020denoising, rombach2022high, dhariwal2021diffusion, saharia2022photorealistic} has significantly enriched capabilities in image generation and editing. Specifically, \cite{tian2023fashion} demonstrates a multi-modal transformer-based architecture that enhances fashion image editing through textual feedback, although it primarily functions within a search model context, contrary to initial interpretations. In our case, we utilize diffusion-based models, specifically InstructPix2pix-Distill and Adobe Firefly Image 3, which allow image editing based on region selection and text description inputs.
\section{Methodology} \label{sec:Methodology}

\subsection{Data}\label{subsec:data}
The dataset used in this study originates from a European fast fashion company and includes real product data. For each product, it provides a front-facing $2098 \times 1500$ pixel image captured under controlled lighting and a uniform white background (see pipeline Figure~\ref{fig:pipeline}) for an example, and a caption which is a composite string of descriptors detailing product design elements, along with 22 categorical features, including product category, fabric, fashion degree, etc., and 13 numerical features, including product cost, product listed price, life cycle, etc. Further details about this dataset can be found in the Appendix. The dataset contains a total of 8,503 products that include the image and caption pair.

\subsection{Caption Text Processing}\label{subsec:caption_text_processing}

\textbf{Caption cleaning} \hspace{0.1cm}
In our dataset, some captions are not formatted perfectly. For example, several captions include unusual non-English symbols, as well as superfluous commas and periods. To address these issues, in the pre-processing caption cleaning step, we first convert all captions to lowercase. Subsequently, we remove extraneous punctuation marks (such as commas and periods) and then divide the caption sentences into lists of short phrases, where each phrase describes one attribute/feature of the product.

\vspace{0.2cm}
\noindent\textbf{Semantic-based Clustering using MinHashLSH}\hspace{0.1cm}
Following the cleaning of captions, each caption is transformed into a list of features. We define the union of all these features as $\bar{\calF}$. During our analysis, we identify multiple groups of features where the elements within each group are synonyms—indicating that they differ in expression but convey highly similar or identical meanings. Examples of such synonym groups include ``cable knit" and ``cable knit fabric", as well as ``hook and zip fastening" and ``zip and hook fastening". Consequently, our set of features, denoted $\bar{\calF}$, is effectively composed of $k$ synonym groups, represented mathematically as:
\begin{equation*}
\bar{\calF} = \bigcup_{i=1}^k \calG_i.
\end{equation*}
To eliminate this redundancy, we have opted to collapse each group $\calG_i$ into a singleton, with only one representative element. Specifically, we select a representative feature $f_i$ from each $\calG_i$ to construct our refined feature set $\calF$, expressed as:
\begin{equation}\label{eq:feature_set}
\calF \bydef \{f_1, f_2, \dots, f_k\}
\end{equation}
This approach ensures a more streamlined and distinct set of features for subsequent analysis.

To achieve this, we implement the following procedure. First, we separate $\bar{\calF}$ into disjoint synonym groups. In this step, it is essential to assess the semantic similarity between features. Our approach entails grouping features whose similarity exceeds a predefined threshold $\tau_0$. Specifically, for any two features $f_i, f_j \in \bar{\calF}$, we define $H(\cdot)$ as our encoding function, which maps each text feature $f_i$ into a $d$-dimensional vector in $\R^d$, where $d$ is a fixed number. We then define $S(\cdot, \cdot)$ as our similarity metric for these vectors. Consequently, we enforce:
\begin{equation*}
\text{if } f_i \in \mathcal{G} \text{ and } S(H(f_i), H(f_j)) > \tau_0, \text{ then } f_j \in \mathcal{G}.
\end{equation*}
This process ensures that features grouped together share a high degree of semantic similarity.

In particular, we employ MinHashLSH (MinHash Locality-Sensitive Hashing) algorithm 
\cite{broder1997resemblance, gionis1999similarity} to achieve efficient clustering. MinHashLSH is a probabilistic method designed to quickly identify approximate similarities between items. The principle underlying this technique is to hash similar items into the same ``bucket" with a high probability, thereby simplifying the task of identifying similar pairs. This is achieved by using multiple MinHash functions that generate compact signatures for sets. These signatures are then stored in a specialized data structure designed to increase the likelihood of collisions (or placement in the same bucket) among similar signatures. This structure makes the clustering and searching for approximate nearest neighbors highly efficient. It is widely used in applications such as document deduplication and clustering of large datasets \cite{tirumala2023d4, Gokaslan2019OpenWeb, bi2024deepseek}.

\vspace{0.2cm}
\noindent \textbf{Representative Feature Selection with LLMs}  \hspace{0.1cm}
For the final stage of our caption processing, it is necessary to select the most representative feature $f_i$ within each synonym group $\mathcal{G}i$. For this task, we utilize large language models, specifically GPT-4 \cite{achiam2023gpt}, to identify the representative terms. Through this process, we successfully converted more than 40 synonym groups $\calG_i$ which contain more than one feature, into singletons with only a single representative feature. Ultimately, we derived a refined set of 1147 distinct features, denoted as $\calF = \{f_1, f_2, \dots, f_{1147}\}$.

In summary, our caption processing procedure is described as follows:
\begin{enumerate}
    \item Clean the captions and aggregate the features into a unified set $\bar{\calF}$.
    \item Utilize MinHash to map feature phrases into fixed-length encoding vectors (we choose $d = 128$).
    \item Apply the Locality Sensitive Hashing (LSH) algorithm to efficiently cluster similar encodings, which correspond to caption phrases with analogous semantic meanings. We set the similarity threshold be $\tau_0=0.8$. This effectively partitions $\bar{\calF}$ into $k$ semantic groups $\calG_1, \calG_2, \dots, \calG_k$.
    \item Query GPT-4 to identify the best representative feature $f_i$ in each group $\calG_i$.  Following this, we construct the final feature set $\calF = \{f_1, f_2, \dots, f_k\}$.
\end{enumerate}

\subsection{Feature Influence Score}
At this stage, the features within $\mathcal{F}$ are well-defined. To determine which features significantly impact fashion design sales and demand, we introduce the ``influence score" $ \textit{Influence}(f_i)$ for each feature $f_i \in \mathcal{F}$. This score is used as a metric to assess the impact of each designed feature on sales. Specifically, we establish a direct correlation between the influence score of each feature and the sales of corresponding fashion products. Initially, we normalize the sales data of all considered fashion products to a range of $[0,1]$ using min-max normalization. Formally, given a set $S$, for each $s \in S$, we define the min-max normalization transformation as 
\[
  N_{S}(s) \bydef  \frac{s - \min(S)}{\max(S) - \min(S)} \in [0,1].
\]
Now let $S$ denote the set of all sales values in our dataset. Then $N_{S}(s)$ denotes the normalized sales in the range between 0 and 1. For a given feature $f_i$, define $S_i$ as the sales for all the products containing this feature in their captions. For the influence score, we naturally average all the sales of the associated products for feature $f_i$, representing the general impact of this feature on the sales:
\begin{equation*}
    \frac{1}{|S_i|} \sum_{s \in S_i}   N_{S}(s) \in [0,1]
\end{equation*}
This average is the primary component in our definition of influence score later.

However, the frequency with which features appear varies significantly. For instance, certain uncommon features may only appear in a single product. In such cases, the influence score is heavily influenced by the sales value of that single product, which introduces a bias due to varying feature frequencies. To mitigate this bias, we incorporate a frequency regularization term into our influence score calculation. We count the frequency of each feature and apply min-max normalization again to these values. Let $P$ be the set of frequencies of features, and use $p_i$ to denote the frequency for feature $f_i$. Eventually, the influence score for each feature $f_i$, with associated product sales $S_i$ and feature frequency $p_i$, is defined as
\begin{equation}\label{eq:influence_score}
    \textit{Influence}(f_i) \bydef \frac{1}{|S_i|} \sum_{s \in S_i} N_S(s) + \lambda \cdot N_P(p_i).
\end{equation}
Here, $\lambda$ is a regularization parameter used to balance the two components of the score. A positive $\lambda$ effectively penalizes features with lower frequencies. Specifically, we set $\lambda = 0.15$ to balance the contributions of both terms within a similar range. Exploration of alternative $\lambda$ values is proposed for future studies. 

Now each feature $f_i$ is assigned a score $\textit{Influence}(f_i)$, which quantifies the quality or popularity of the feature. We can rank the features in descending order based on their influence scores. This ranking allows designers to prioritize the inclusion of ``good" features—those at the top of the list—in their fashion designs, while steering clear of ``bad" features, which are positioned at the bottom. This strategic selection aims to enhance the appeal and marketability of the resulting fashion products.

\subsection{Forecasting Model}

To assess the potential market performance of fashion products, we need to develop a model capable of evaluating product popularity based on available data. In this study, we assume that a product's sales is positively correlated with its popularity \cite{moen2017online}. This assumption allows us to treat sales as a proxy for popularity, making it possible to assess the relative appeal of a product. In this section, we introduce our forecasting model, which we call \textit{Fashion Demand Predictor (FDP)}, which can be used to classify products into different sales categories based on multi-modal inputs, including the image and some basic information (such as product names and types) of the products.

\subsubsection{Fashion Demand Predictor (FDP)}
\vspace{0.2 cm}
\textbf{Sales Classification Task}\hspace{0.1 cm}
Our Fashion Demand Predictor (FDP) is designed to assess the market viability of new fashion designs. Prior research\cite{sha2024imaged} has explored the use of various machine learning algorithms to predict the sales of individual fashion items. Their studies underscore the difficulty of precisely forecasting exact sales figures for each product. However, for the purposes of our study, determining exact sales numbers is less critical than assessing relative popularity among products. To simplify this task, we convert the continuous sales data into categorical classes. This effectively transforms the challenging task of exact sales forecasting into a more manageable classification problem, simplifying the prediction process.

\begin{figure}[ht]
\begin{center}
    \includegraphics[width=0.48\textwidth]{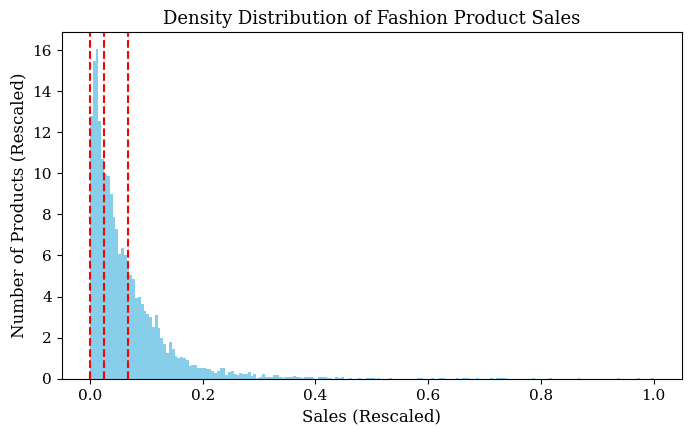} 
\end{center}
\caption{\small Density distribution of normalized fashion product sales. The sales are normalized between 0 and 1 using min-max scaling, showing a skewed distribution with a higher concentration of lower sales values. The red dashed lines indicate thresholds that divide the data into three equal quantiles. }
\label{fig:density}
\end{figure}

\vspace{0.2 cm}
\noindent \textbf{Sales Distribution Study} \hspace{0.1 cm}
To determine an appropriate sales class label for each product, we first study the total sales distribution for all the products. As shown in Figure~\ref{fig:density}, the distribution is heavily right-skewed with a long tail to the right-hand side, indicating very few products achieve extremely high sales. In this study, we are interested in the relative popularity of modified new products compared to the original design. This supports the use of the equal quantiles method to generate sales class labels, rather than using fixed sales value intervals. Specifically, we divide the distribution of sales into three regions of equal probability (see Figure~\ref{fig:density}), and assign the class labels 1, 2, and 3 respectively, where a higher label indicates higher sales. By using equal quantiles, each class contains the same number of products, ensuring balanced representation across classes. This is particularly beneficial for our classification model, as class imbalance could introduce additional challenges to the learning process\cite{johnson2019survey}.

\vspace{0.2 cm}
\noindent \textbf{Objective Function of FDP} \hspace{0.1cm} Different types of fashion product features are described in details found in the Appendix. As mentioned in the previous sections, in this work, we aim to solve the following objective. Given 36 categorical and numeric product data features denoted as $X_i$, a text caption denoted as $t_i$, an image embedding feature denoted as $e_i$, we want to predict the correct sales class label. The correct sales class label, denoted as $y_i$, is derived by dividing the continuous true sales data into k quantile-based classes. We can formulate the objective of our FDP as follows:
\begin{equation}
    \begin{aligned}
    \underset{ f}{\text{minimize}} \quad & \frac{\sum_{i=1}^{n} \mathbb{I}(y_i \neq \hat{y_i})}{n} \\
    \text{where} \quad & \hat{y_i} = f(X_i, t_i, e_i)
    \end{aligned}
\end{equation}

\noindent The objective here is to identify the optimal classifier $f$ to minimize the average classification error, or in other words, to improve the classification accuracy. 

\begin{figure*}[ht]
    \centering
    \includegraphics[width=\textwidth]{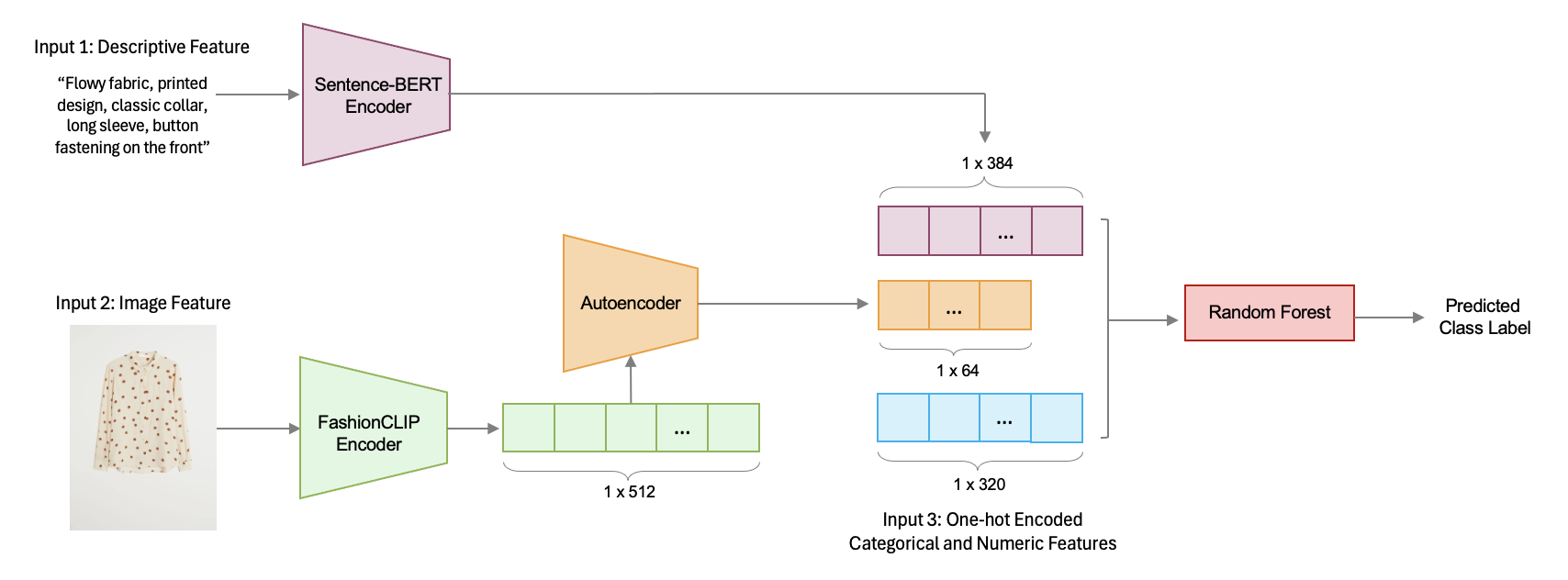}  %
    \caption{A schematic overview o Fashion Demand Predictor(FDP): the model takes three types of input features: (1) descriptive text features processed by the Sentence-BERT's encoder, (2) image features processed by the FashionCLIP's encoder, and (3) one-hot encoded categorical and numeric features. The encoded descriptive and image features are further compressed by an autoencoder.}
    \label{fig:FDP}
\end{figure*}

\vspace{0.2 cm}
\noindent \textbf{Data Preprocessing} \hspace{0.1 cm} Some preprocessing techniques are used to ensure that: (1) the machine learning model can correctly handle unstructured data types such as text and images, and (2) the features have manageable dimensionality so that no single feature dominates. All the categorical features are one-hot encoded to maintain an equal-weighted representation for each distinct category. To handle the caption, we use the Sentence-BERT \cite{reimers2019sentence} to encode the string of descriptors into dense vector representations. For the images embeddings, we use FashionCLIP \cite{chia2022fashionclip}, an open-source pre-trained transformer model designed for fashion industry applications. Due to the high dimensionality of the image embeddings, we implement dimensionality reduction using an autoencoder to preserve only the most essential features. The classification task using FDP is illustrated in Figure~\ref{fig:FDP}. The train and test sets were created using a random split with an 4:1 ratio, where 80\% of the data was used for training and 20\% for testing.

\vspace{0.2 cm}
\noindent\textbf{FDP Architecture and Accuracy} \hspace{0.1cm} 
We have tried several classification models, including Support Vector Machines (SVM), Random Forest, and XGBoost, among others. We opted for these lighter, simpler machine learning models because they are cost-effective and efficient. Among these classifiers, the Random Forest model has demonstrated the best performance. As an ensemble learning method, the Random Forest classifier constructs multiple decision trees and aggregates their predictions \cite{parmar2019review}. Its resilience to noisy data and capability to manage high-dimensional datasets—like ours, which contains over 768 features following data preprocessing—make it particularly effective for our classification task\cite{parmar2019review}.

Note that instead of getting the final classification class, by taking out the logits, we are able to obtain a probability distribution over all possible classes after the softmax function. By analyzing these probabilities later, we can determine if one product is more popular than another, as a small shift in sales may not alter the predicted class, but the underlying probability distribution provides more nuanced insights. In Section~\ref{subsec:experiment_setup}, we discuss in detail how we transform the per-class probability into an overall score that can be used to compare relative product popularity.

In Table~\ref{tab:FDP_acccuracy}, the classification accuracy of the Random Forest FDP model is shown. Moreover, we demonstrated the 
accuracies under different numbers of sales classes (denoted as $C$). Intuitively, the higher the number of classes, the more challenging it is for FDP to achieve a satisfactory accuracy. When we take $C = 3$, the classification accuracy of Random Forest exceeds 80\%. Thus, for the rest of this paper, we assign all products into three distinct sales classes, i.e., low-sales class, medium-sales class, and high-sales class.

\begin{table}[h]
\centering
\setlength{\tabcolsep}{10pt} 
\begin{tabular}{|c|c|c|c|}
\hline
Number of classes $C$ & 3 & 4 & 5 \\ \hline
Classification accuracy & \textbf{0.81} & 0.71 & 0.61 \\ \hline
\end{tabular}
\caption{Prediction accuracy of FDP on the test dataset across different choices of the number of classes. 
\label{tab:FDP_acccuracy}}
\end{table}


\section{New Product Design} \label{sec:Design}

After identifying the ``good" and ``bad" features, our next step is to generate new fashion designs and use ablation studies to assess how these features affect product quality. We experiment with several diffusion models for image editing to modify the original designs. Ultimately, two models emerge as particularly effective, providing relatively good results in editing. The first tool we apply is called \textit{InstructPix2pix-Distill} \cite{instruct_pix2pix_distill}, a distilled version of \textit{InstructPix2Pix}, which is a pre-trained text-guided image editing tool developed by \cite{brooks2023instructpix2pix}, which allows for precise modifications based on natural language inputs. \textit{InstructPix2pix-Distill} improves the efficiency by incorporating distillation techniques to speed up the performance. This algorithm modifies the image as a whole, without allowing us to specify particular regions for editing. While this limits control over localized changes, it also preserves the overall image integrity, ensuring that the edits do not disrupt the cohesion of the original design. One caveat of this model is that it is not fine-tuned for fashion products, which led to some difficulties in understanding certain fashion-specific terminologies. Another model we use is the \textit{Adobe Firefly Image 3 Model} \cite{adobe2024firefly}. \textit{Firefly} is a trained generative AI model developed by Adobe, offering advanced image generation and editing capabilities. Unlike  \textit{InstructPix2pix-Distill}, it allows for more precise control by enabling region-specific modifications, where users can specify areas of an image to edit, giving us the flexibility to focus on particular parts of the design. However, one drawback is that it can generate content that does not fit well with the rest of the product, leading to inconsistencies in style. Despite this, Firefly excels in producing highly detailed and photorealistic outputs. Its ability to allow region-specific edits made it a great complement to \textit{InstructPix2pix-Distill} in our research.
\section{Experiments} \label{sec:Experiments}

\subsection{Experiment Setup and Evaluations}\label{subsec:experiment_setup}

While models can readily provide forecast probabilities for a product's likelihood of belonging to various sales classes, this task is highly challenging for humans to perform accurately. Our objective is not to predict exact product sales, but rather to compare the relative popularity of a product before and after modifications. To achieve this, we transform the model's predicted class probabilities into an overall prediction score, $s_j$, for product $j$, using the following equation:
\begin{equation}\label{eq:FDP_averaged_class}
    s_j \bydef \sum_{i=1}^{n} \mathbb{P}(C_{ji}) \cdot i, \quad i \in \{1,2,3\}
\end{equation} 

\noindent where $\mathbb{P}(C_{ji})$ represents the probability of product $j$ belonging to the $i$-th class. Using this approach, we design two experiments below to evaluate our model's performance.

\vspace{0.2 cm}
\noindent\textbf{Experiment 1: Image Triplets Ranking} \hspace{0.1cm}
The first experiment assesses how effectively the model ranks the popularity of products within the same category. The second experiment compares the original product images with modified versions generated by diffusion models, aiming to see if the predicted popularity aligns with our expectations for increased or decreased appeal. In the first experiment, we selected $m=20$ triples $T_1, T_2, \dots, T_m$, where in each triple, ensuring that each triple contains three products of the same type, each with distinct sales class labels of 1, 2, and 3, respectively. We evaluate the reliability of our model, FDP, by calculating the Kendall Tau score \cite{kendall1938new} to see how well its internal ranking aligns with the ground truth, represented by the sales class label for each product. We define the total Kendall Tau score as 
\begin{equation}
\tau(T_1, T_2, \dots, T_m) = \sum_{i=1}^{m} \frac{P_i - Q_i}{\binom{n}{2}}
\end{equation}
Here $P_i$ represents the number of concordant pairs for the $i$-th triple, and $Q_i$ represents the number of discordant pairs for the same triple, compared against the ground truth ranking (more details can be found in \cite{kendall1938new}). The term $\binom{n}{2}$ is the total number of possible pairs in a triple, where $n = 3$. The value $\tau$ is in the range of $[-1, 1]$, where a higher value indicates better alignment. For the baseline, we prompt \textit{Llava-v1.5-7b} \cite{liu2024visual}, one of the leading open-source multi-modal models available that support both image and text inputs, to rank the three images and calculate its Kendall Tau score similarly.

Moreover, instead of only using the ground truth class labels, we further validate the performance of our FDP by comparing it against the rankings of human raters. We recruited 100 female survey participants through \textit{Prolific}, an online survey platform. Detailed demographic information for the participants is provided in the Appendix. We chose female participants specifically because our fashion products are designed and marketed for a female audience, ensuring that the feedback aligns with the target consumer demographic.

\noindent\textbf{Experiment 2: Ablation Case Study} \hspace{0.1cm} In the second experiment, we select 9 ``good" features and 9 ``bad" features according to their influence score defined in \ref{eq:influence_score}. For each feature, we select one associated product and conduct an ablation by removing that feature from the original product. Then and evaluate how the modification affects the product's popularity prediction of our FDP model, described in 
\ref{eq:FDP_averaged_class}. Ideally, removing ``good" features should result in a decrease in product popularity, while removing ``bad" features is expected to increase it. To verify this assumption, we asked human raters to indicate their preferences between the original image and the AI-modified image.

During the feature selection process, we adhered to two criteria: (1) the features must be easily editable using the diffusion models described in Section~\ref{sec:Design}, and (2) the features must be visually identifiable to human observers. For instance, features associated with specific parts of clothing, such as logo prints or pockets, are simpler to modify compared to features that describe the texture or fabric, which may pose greater challenges for visual editing.

\subsection{Results} \label{subsec:Results}
\vspace{0.2cm}
\noindent \textbf{Results for Experiment 1}\hspace{0.1cm} 
The results from our triplet product ranking task (Experiment 1) are summarized in Table~\ref{tab:experiment1_results}.

\begin{table}[h]
\centering
\small
\begin{minipage}{0.45\textwidth} 
    \centering
    \begin{tabular}{|c|c|c|}
    \hline
    & \textbf{Class Labels} & \textbf{Human Survey} \\ \hline
    \textbf{FDP} & 0.93  & 0.43 \\ \hline
    \textbf{Llava} & -0.56  & -0.30  \\ \hline
    \end{tabular}
    \subcaption{Kendall Tau Score}
    \label{tab:kendall_tau_scores}
\end{minipage}%

\hspace{1cm} 
\begin{minipage}{0.45\textwidth}
    \centering
    \begin{tabular}{|c|c|c|}
    \hline & \textbf{Class Labels} & \textbf{Human Survey} \\ \hline
    \textbf{FDP} &  18 & 8 \\ \hline
    \textbf{Llava} &  1 & 1 \\ \hline
    \end{tabular}
    \subcaption{Triplet Correct Counts}
    \label{tab:triplet_correct_counts}
\end{minipage}
\caption{Comparison between FDP and Llava on Class Labels and Human Survey}
\label{tab:experiment1_results}
\end{table}
\noindent As previously mentioned, for this experiment, we consider two types of ``ground truth": one derived from sales class labels and the other from relative rankings collected from the human survey. The sales class label provides an objective baseline based on actual product performance, capturing real-world purchasing trends. In contrast, the human survey ranking reflects subjective preferences based solely on the product's image and its listed price, without access to additional product details or the ability to physically interact with the items, such as trying them on. We calculated the Kendall Tau score between these two ground truths and obtained a value of 0.53. This score confirms that, despite some discrepancies, the two ground truths are reasonably well-aligned and positively correlated.

Based on the results in Table~\ref{tab:experiment1_results}, the performance of the Fashion Demand Predictor (FDP) model significantly surpasses that of the Llava model in both alignment with class labels and with human survey rankings. In Table~\ref{tab:kendall_tau_scores}, the Kendall Tau scores indicate a high correlation between FDP predictions and the class labels, as well as a positive correlation with human survey rankings. These results suggest that FDP captures trends that are consistent with actual sales data and, to a lesser extent, with subjective human preferences. Conversely, the Llava model shows negative correlations with both class labels and human survey rankings, indicating that its predictions diverge significantly from both real-world performance and human preferences.  In Table~\ref{tab:triplet_correct_counts}, we further counted the number of triples where the two rankings match perfectly. The results further reinforces FDP's superiority, showing that FDP achieves a high count of correct triplet rankings compared to class labels and human survey rankings. In contrast, Llava achieves only one correct triplet ranking against each baseline, suggesting that it fails to reliably rank products in a manner consistent with actual performance or consumer preferences. Overall, these results highlight FDP's effectiveness in aligning with both objective sales data and human preferences.

\begin{table}[ht]
\centering
\setlength{\tabcolsep}{3.0pt}
\fontsize{8pt}{10pt}\selectfont
\renewcommand{\arraystretch}{1.5}
\begin{tabular}{|c|p{2.3cm}|c|c|c|}  
\hline
\textbf{Remove} & \textbf{Feature Name} & \textbf{Feature Score} & \textbf{Original} & \textbf{AI-Modified} \\
\hline
\multirow{9}{*}{Good} 
& Folded Cuffs  & 0.258 & 2.636 \text{$\uparrow$} & 2.440 \\
\cline{2-5}
& Three Zipped Pockets & 0.210 & 2.566 \text{$\uparrow$} & 2.448 \\
\cline{2-5}
& Short Raglan Sleeve & 0.228 & 2.925 \text{$\uparrow$} & 2.844 \\
\cline{2-5}
& Adjustable Inner Drawstring Waist (Green) & 0.232 & 2.616 \text{$\uparrow$} & 2.507 \\
\cline{2-5}
& Adjustable Inner Drawstring Waist (Yellow) & 0.232 & 2.636 \text{$\uparrow$} & 2.502 \\
\cline{2-5}
& Light Vintage Wash & 0.307 & 2.593 \text{$\uparrow$} & 2.434 \\
\cline{2-5}
& Palm Print (Tree) & 0.160 & 2.521 \text{$\uparrow$} & 2.459 \\
\cline{2-5}
& Palm Print (Leaf) & 0.160 & 2.450 \text{$\uparrow$} & 2.380 \\
\cline{2-5}
& Multicolour Print & 0.171 & 2.761 \text{$\uparrow$} & 2.693 \\
\hline
\multirow{9}{*}{Bad} 
& Contrasting Appliques & 3.39E-04 & 1.157 & 1.249 \text{$\uparrow$} \\
\cline{2-5}
& Notched Lapel & 6.40E-04 & 1.228 & 1.310 \text{$\uparrow$} \\
\cline{2-5}
& Zip Detail & 2.14E-03 & 1.173 & 1.215 \text{$\uparrow$} \\
\cline{2-5}
& Contrasting Pocket & 3.02E-03 & 1.223 & 1.282 \text{$\uparrow$} \\
\cline{2-5}
& Decorative Pleats & 2.68E-03 & 1.162 & 1.202 \text{$\uparrow$} \\
\cline{2-5}
& Glass Beads & 4.68E-03 & 1.387 & 1.418 \text{$\uparrow$} \\
\cline{2-5}
& Lapels with Press Studs & 3.30E-03 & 1.310 & 1.417 \text{$\uparrow$} \\
\cline{2-5}
& Stripped Finish & 2.95E-04 & 1.345 \text{$\uparrow$} & 1.342 \\
\cline{2-5}
& Leather Inner Lining & 1.90E-03 & 1.419 \text{$\uparrow$} & 1.397 \\
\hline
\end{tabular}
\caption{Comparison of the FDP prediction scores between original and AI-modified designs after removing ``good" and ``bad" Features. \text{$\uparrow$} indicates the design with a higher score.}
\label{tab:FDP_prob_changes}
\end{table}

\noindent \textbf{Results for Experiment 2}\hspace{0.1 cm}
In Table~\ref{tab:FDP_prob_changes}, we compare the FDP prediction scores for original and AI-modified designs after removing identified ``good" and ``bad" features. For ``good" features, the original designs consistently receive higher popularity scores than the AI-modified versions. This drop in popularity for AI-modified designs after the removal of these ``good" features, aligns with our expectations, confirming that these features positively impact the appeal of the product. Conversely, for ``bad" features, the AI-modified designs score higher than the original designs in 7 out of 9 cases, indicating that FDP supports our hypothesis that removing these features enhances product appeal.

\begin{table}[ht]
\centering
\setlength{\tabcolsep}{4pt} 
\fontsize{8pt}{8pt}\selectfont
\renewcommand{\arraystretch}{1.5} 
\begin{tabular}{|c|p{2.3cm}|c|c|} 
\hline
\textbf{Remove} & \textbf{Feature Name} & \textbf{Original (\%)} & \textbf{AI-Assisted (\%)} \\
\hline
Good & Folded Cuffs & 53.6 $\uparrow$ & 46.4 \\
\hline
Good & Three Zipped Pockets & 92.9 $\uparrow$ & 7.1 \\
\hline
Good & Short Raglan Sleeve & 41.1 & 58.9 $\uparrow$ \\
\hline
Good & Adjustable Inner Drawstring Waist (Yellow) & 55.4 $\uparrow$ & 44.6 \\
\hline
Good & Adjustable Inner Drawstring Waist (Green) & 62.5 $\uparrow$ & 37.5 \\
\hline
Good & Light Vintage Wash & 42.9 & 57.1 $\uparrow$ \\
\hline
Good & Palm Print (Tree) & 70.5 $\uparrow$ & 29.5 \\
\hline
Good & Palm Print (Leaf) & 80.4 $\uparrow$ & 19.6 \\
\hline
Good & Multicolour Print & 52.7 $\uparrow$ & 47.3 \\
\hline
Bad & Contrasting Appliqués & 75.9 $\uparrow$ & 24.1 \\
\hline
Bad & Notched Lapel & 46.4 & 53.6 $\uparrow$ \\
\hline
Bad & Zip Detail & 46.4 & 53.6 $\uparrow$ \\
\hline
Bad & Contrasting Pocket & 74.1 $\uparrow$ & 25.9 \\
\hline
Bad & Decorative Pleats & 50 & 50 \\
\hline
Bad & Glass Beads & 44.6 & 55.4 $\uparrow$ \\
\hline
Bad & Lapels with Press Studs & 83 $\uparrow$ & 17 \\
\hline
Bad & Stripped Finish & 46.4 & 53.6 $\uparrow$ \\
\hline
Bad & Leather Inner Lining & 43.8 & 56.2 $\uparrow$ \\
\hline
\end{tabular}
\caption{Comparison of human survey preference scores between original and AI-modified designs after removing ``good" and ``bad" Features. $\uparrow$ indicates the design with a higher preference score.}
\label{tab:human_product_percentage}
\end{table}

The survey results, displayed in Table~\ref{tab:human_product_percentage}, provide an interesting comparison to the FDP model's predictions. For designs with ``good" features removed, the survey participants, like the FDP model, overwhelmingly preferred the original versions, reinforcing our hypothesis that removing these high-influence score features diminishes the product's appeal. For the removal of ``bad" features, the survey results show a slightly more varied response. Human participants favor the AI-modified designs in 5 out of 9 cases, prefer the original design in 3 cases, and show no clear preference in 1 case. Although the survey results are more diverse than the FDP model’s predictions, they still generally support the conclusion that removing ``bad" features can enhance product appeal. This nuanced divergence indicates that, while the removal of low-influence features often improves the design, it does not universally lead to a more attractive product. These findings reflect the complexity of feature influence in fashion design.

\section{Discussion} \label{sec:Discussion}
In this work, we introduce a comprehensive, end-to-end framework that identifies the most influential design features in fashion images and guides both design modifications and new product evaluations. Our methodology leverages the newly proposed influence score and the Fashion Demand Predictor (FDP) model, which aligns closely with human raters’ assessments of product popularity. There are some limitations to our current work, including the need to adjust the regularization parameter $\lambda$, the challenge of conducting surveys with larger populations due to high costs, and the limitation to only removing features rather than also adding or replacing features (mostly limited by the current capabilities of AI models in image editing). Despite these challenges, our framework represents, to the best of our knowledge, the first fully automated and systematic approach to fashion image analysis that facilitates in-depth product reviews without the need for expert inspection.

{
    \small
    \bibliographystyle{ieeenat_fullname}
    \bibliography{main}
}

\newpage

\section{Appendix}

\subsection{Details of Data}

This section provides additional context on the dataset and survey used in this study. Table~\ref{tab:data description} lists all relevant columns (13 numerical features and 22 categorical features) from the dataset, outlining the features that contributed to the analysis and modeling process. Tables~\ref{tab:survey age} and \ref{tab:region_distribution} present the demographic distribution of survey participants, offering insights into the age range and diversity of respondents for the Image Triplet Ranking and Ablation Case Study tasks. The regional information is only available for the Image Triplet Ranking Task because the two surveys were designed using different tools, and only the tool (Qualtrics) used for the Triplet task recorded participants' latitude and longitude data.

\begin{table}[h!]
\centering
\begin{tabular}{|c|c|}
\hline
\textbf{Numerical Features} & \textbf{Categorical Features} \\ \hline
Listed Price & Product Family \\ \cline{1-2}
Product Cost & Product Category \\ \cline{1-2}
Lifecycle & Product Specific Type \\ \cline{1-2}
Number of Sizes & Color Spectrum \\ \cline{1-2}
Cosine Transformation of Dates & Color Family \\ \cline{1-2}
Sine Transformation of Dates & Fabric \\ \cline{1-2}
Same Family Product Count & Fashion Degree \\ \cline{1-2}
\begin{tabular}[t]{@{}c@{}}Same Family, Same Color \\ Product Count\end{tabular} & \begin{tabular}[c]{@{}c@{}}Special Style \\ Binary Indicator\end{tabular} \\ \cline{1-2}
\begin{tabular}[t]{@{}c@{}}Same Family, Same Price \\ Product Count\end{tabular} & Length Specification \\ \cline{1-2}
\begin{tabular}[t]{@{}c@{}}Same Family, Same Price, \\ Same Color Product Count\end{tabular} & Neck Design \\ \cline{1-2}
Family Price Minimum & Structure Type \\ \cline{1-2}
Family Price Median & Knit Style \\ \cline{1-2}
Family Price Maximum & Fit Style \\ \cline{1-2}
& Pattern Inclusion \\ \cline{1-2}
& Sleeve Length \\ \cline{1-2}
& Release Month \\ \cline{1-2}
& Release Year \\ \cline{1-2}
& Season Code \\ \cline{1-2}
& Special Style Category \\ \cline{1-2}
& Store Type \\ \hline 
& \begin{tabular}[c]{@{}c@{}}During Christmas \\ Binary Indicator\end{tabular} \\ \hline
& \begin{tabular}[c]{@{}c@{}}During Black Friday \\ Binary Indicator\end{tabular} \\ \hline
\end{tabular}
\caption{List of Numerical and Categorical Features in Our Data}
\label{tab:data description}
\end{table}

\begin{table}[H]
\centering
\setlength{\tabcolsep}{3pt} 
\begin{tabular}{|c|c|c|}
\hline
\textbf{Age Group} & \textbf{Image Triplet Ranking} & \textbf{Ablation Case Study} \\ \hline
18-25 & 39 (39.0\%) & 34 (30.4\%) \\ \hline
26-35 & 36 (36.0\%) & 62 (55.4\%) \\ \hline
35+ & 25 (25.0\%) & 16 (14.3\%) \\ \hline
Total & 100 & 112 \\ \hline
\end{tabular}
\caption{Age Distribution of Survey Takers for Image Triplet Ranking and Ablation Case Study Tasks}
\label{tab:survey age}
\end{table}

\begin{table}[H]
\centering
\setlength{\tabcolsep}{24pt} 
\begin{tabular}{|l|c|}
\hline
\textbf{Country}       & \textbf{Percentage (\%)} \\ \hline
South Africa           & 30.0                     \\ \hline
United Kingdom         & 17.0                     \\ \hline
Canada                 & 8.0                      \\ \hline
Portugal               & 8.0                      \\ \hline
United States          & 8.0                      \\ \hline
Poland                 & 6.0                      \\ \hline
Mexico                 & 3.0                      \\ \hline
Hungary                & 3.0                      \\ \hline
Others       & 17.0
           \\ \hline
\end{tabular}
\caption{Distribution of Regions of Survey Takers for Image Triplet Ranking Task by Percentage. Countries in "Others" include Spain, Greece, Italy, Czech Republic, Israel, Germany, Belgium, Brazil, Sweden, Morocco, India, Switzerland, and Kenya.}
\label{tab:region_distribution}
\end{table}

\subsection{Examples of Images}

\begin{figure}[H]
\centering
\includegraphics[width=0.45\textwidth]{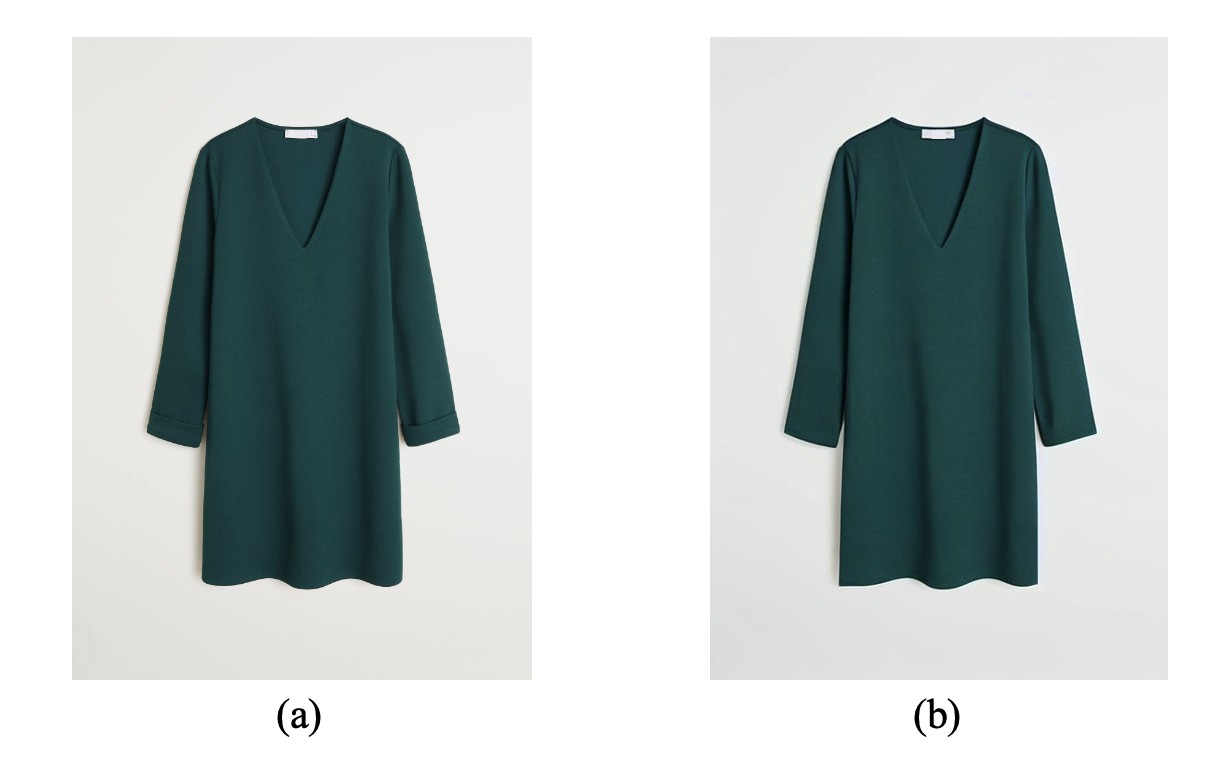} 
\caption{Comparison of Original (left) and AI-Modified (right) Designs. The AI-Modified design demonstrates the effect of removing a ``Good'' feature \textit{Folded Cuffs}.}
\label{fig:combined_image_example1}
\end{figure}

\begin{figure}[H]
\centering
\includegraphics[width=0.45\textwidth]{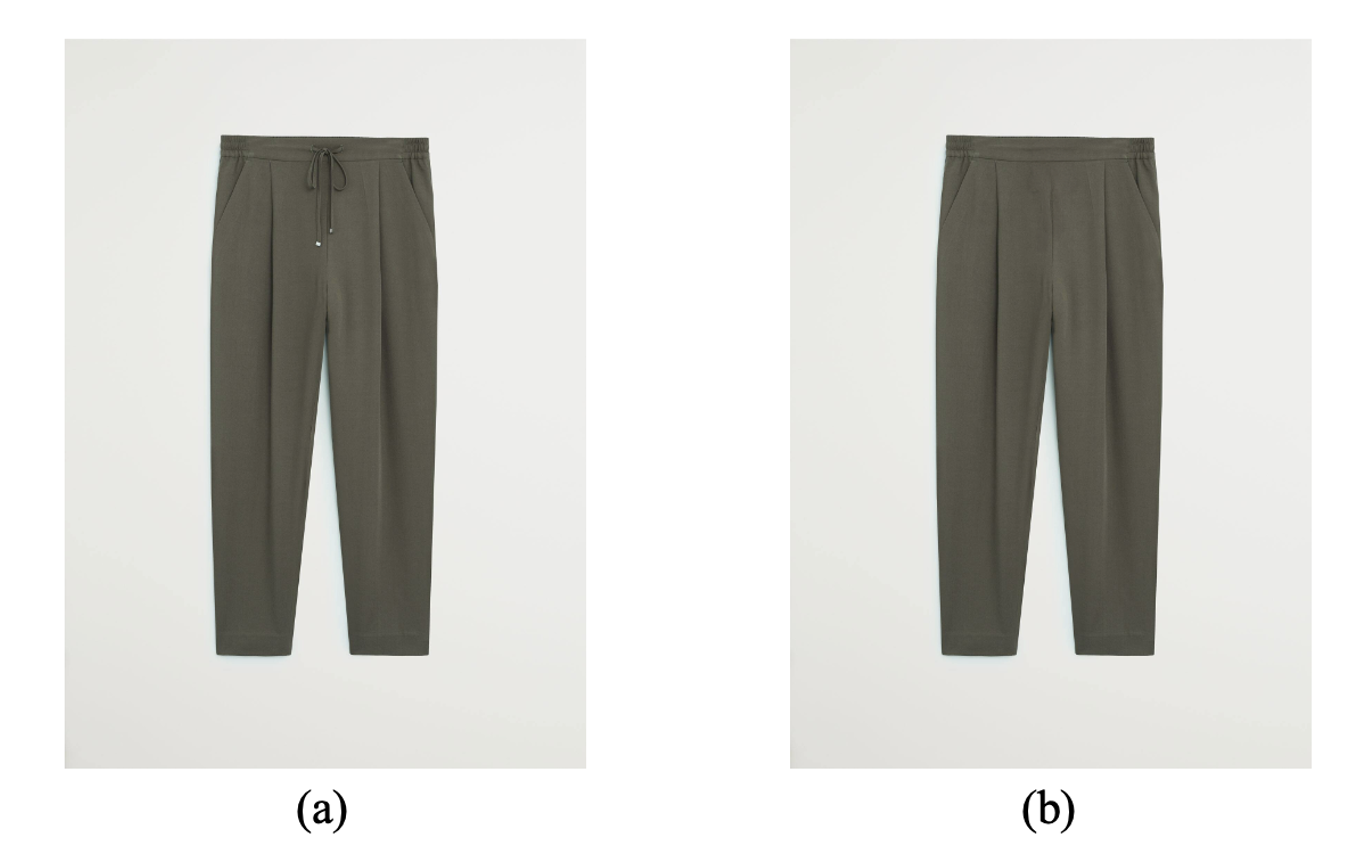} 
\caption{Comparison of Original (left) and AI-Modified (right) Designs. The AI-Modified design demonstrates the effect of removing a ``Good'' feature \textit{Adjustable Inner Drawstring Waist}.}
\label{fig:combined_image_example2}
\end{figure}

\begin{figure}[H]
\centering
\includegraphics[width=0.5\textwidth]{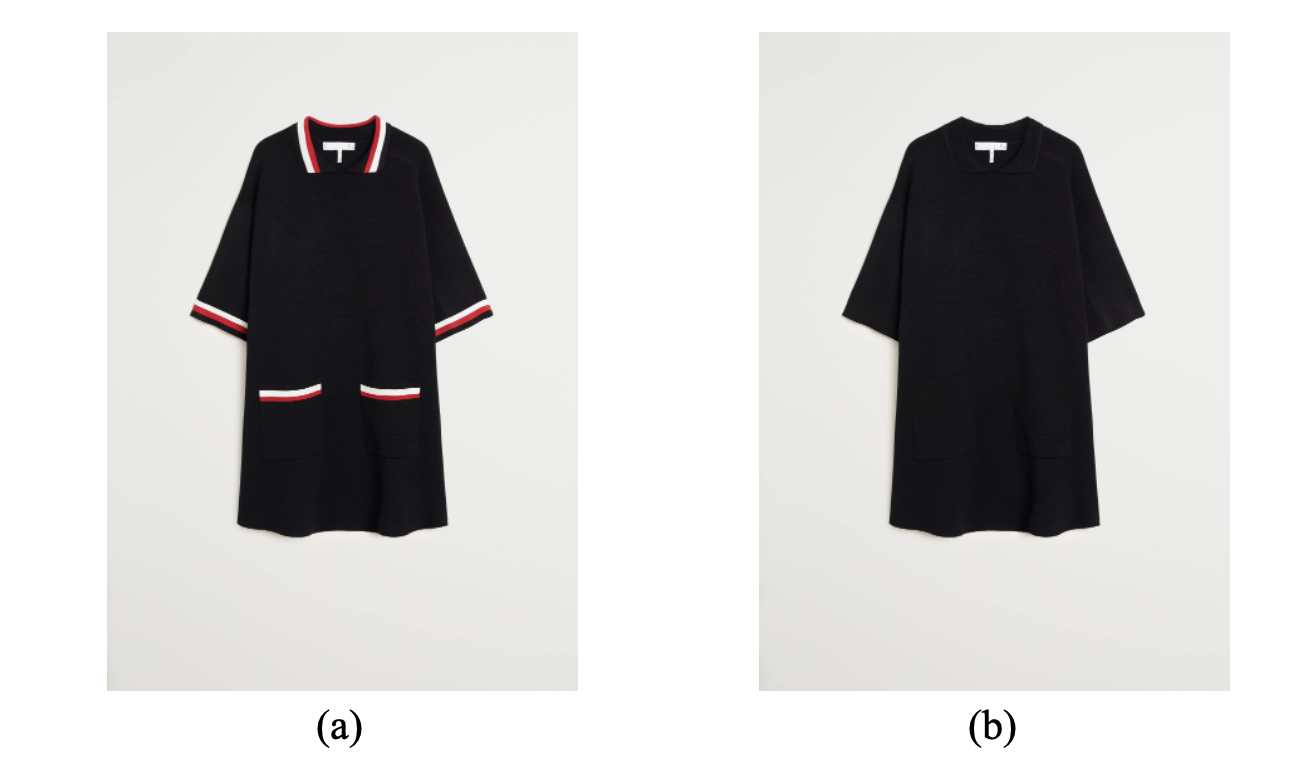} 
\caption{Comparison of Original (left) and AI-Modified (right) Designs. The AI-Modified design demonstrates the effect of removing a ``Bad'' feature \textit{striped pattern}.}
\label{fig:combined_image_example3}
\end{figure}

In this section, we present three examples of our image modification process. Each pair of images compares the original design (left) with an AI-modified design (right), demonstrating the effect of removing specific features. These examples showcase how the removal of ``good" or ``bad" features influences the overall aesthetics and functionality of the designs. In Figure~\ref{fig:combined_image_example1}
, the AI-modified design removes the cuff detail, a feature that contributes to the garment's structured look. Figure~\ref{fig:combined_image_example2} showcases the removal of the inner drawstring at the waist, which is a functional feature providing adjustability and customization. Finally, Figure~\ref{fig:combined_image_example3} illustrates the elimination of the striped pattern, simplifying the design for a cleaner and more minimalistic aesthetic.

\subsection{Prompt for Llava}
In Section~\ref{subsec:Results}, we utilized Llava-1.5  for zero-shot inference to generate the ranking of product triplets, serving as our baseline. Specifically, we generated a single image containing three product images from each triplet and then queried Llava to produce popularity scores for these products (the ranking is generated later by comparing the scores). This approach was adopted because direct prompts to Llava for generating rankings resulted in highly biased and unstable outcomes; notably, the rankings often remained unchanged even when the order of the products was altered. This method is closely related to common applications of LLMs, particularly in settings of ``LLM-as-a-judge", where in our case the LLM judges the popularity as the quality \cite{huang2024empirical, li2023alpacaeval, li2024rule, zhong2023agieval, leerlaif}. The prompt template used for querying Llava, which includes the image along with basic product information such as price and category, is detailed in Table~\ref{tab:llava_rating_prompt}.
\begin{figure}[H]
\centering
\small
\begin{tcolorbox}[colback=cyan!10!white, 
                  colframe=cyan!30!white, 
                  width=\columnwidth, 
                  arc=4mm, 
                  auto outer arc,
                  ]
This image features three fashion design products, labeled from left to right as A, B, and C. Below are the details of the three products:
\\
\texttt{\color{red}<BASIC\_INFORMATION>}
\\
\\
Considering both the design quality and their information, please estimate the popularity of each product and provide an estimated popularity score on a scale from 0 to 100 for each product. Make sure the three scores are different. Reply with three numbers.

\end{tcolorbox}
\caption{Llava prompt for generating the popularity scores.}
\label{tab:llava_rating_prompt}
\end{figure}

\subsection{Fine-tuning Llava}
We also experimented with fine-tuning Llava for a classification task involving images labeled by sales class, followed by its application in triplet ranking as discussed in Section~\ref{subsec:Results}. However, we observed that it is difficult for Llava to accurately learn the sales class; after fine-tuning, Llava’s predictions yielded highly unbalanced class labels compared to the original distribution of the sales classes, suggesting possible overfitting. Furthermore, it is important to note the significant computational costs associated with fine-tuning Llava. In contrast to our lightweight forecasting models using Random Forest, which only require 15 minutes on a CPU to train with a dataset of 8503 product images, fine-tuning Llava—even in its 8-bit quantized version—requires several GPU hours using an NVIDIA A100-80GB with a LoRA rank of 64. Indeed, even zero-shot inference using Llava is resource-intensive. Consequently, our method is not only more cost-effective but also demonstrates higher accuracy in our experiments.

\end{document}